\documentclass[10pt,twocolumn,letterpaper]{article}

\usepackage[pagenumbers]{iccv} 
\usepackage{microtype}
\usepackage{graphicx}
\usepackage{booktabs} 
\usepackage{hyperref}
\usepackage{amsmath}
\usepackage{caption}
\usepackage{pifont}
\usepackage{float}
\usepackage{multirow}
\usepackage{amssymb}
\usepackage{amsthm}
\usepackage{comment}
\usepackage{colortbl}
\usepackage{xcolor}

\usepackage{subcaption}
\expandafter\def\csname ver@subfig.sty\endcsname{}
\usepackage{svg}
\usepackage{hyperref}

\definecolor{iccvblue}{rgb}{0.21,0.49,0.74}

\title{DeskVision: Large Scale Desktop Region Captioning for Advanced GUI Agents}

\author{
	Yibin Xu\thanks{Equal contribution.}\quad  
	Liang Yang\footnotemark[1] \quad
	Hao Chen\footnotemark[1]
	 \\
	Hua Wang\quad
    Zhi Chen \quad
    Yaohua Tang\thanks{\texttt{tangyaohua28@gmail.com}}
	\\\\
	Moore Threads AI
	\\
}

\begin{document}
\maketitle
\begin{abstract}
The limitation of graphical user interface (GUI) data has been a significant barrier to the development of GUI agents today, especially for the desktop / computer use scenarios. To address this, we propose an automated GUI data generation pipeline, AutoCaptioner, which generates data with rich descriptions while minimizing human effort. Using AutoCaptioner, we created a novel large-scale desktop GUI dataset, DeskVision, along with the largest desktop test benchmark, DeskVision-Eval, which reflects daily usage and covers diverse systems and UI elements, each with rich descriptions. With DeskVision, we train a new GUI understanding model, GUIExplorer. Results show that GUIExplorer achieves state-of-the-art (SOTA) performance in understanding/grounding visual elements without the need for complex architectural designs. We further validated the effectiveness of the DeskVision dataset through ablation studies on various large visual language models (LVLMs). We believe that AutoCaptioner and DeskVision  will significantly advance the development of GUI agents, and will open-source them for the community. 
\end{abstract}
\section{Introduction}
\label{sec:intro}


The graphical user interface (GUI) has become an essential medium for human interaction with the digital world, facilitating everything from everyday tasks—such as sending emails or booking flights—to complex professional operations. 
The emergence of Large Language Models (LLMs) \cite{dubey2024llama, achiam2023gpt} and Large Vision-Language Models (LVLMs) \cite{liu2024visual, zhu2023minigpt, ye2024mplug} has fueled growing interest in developing intelligent agent systems capable of understanding and operating GUIs. These agents can serve as virtual assistants, automating tasks, streamlining workflows, and, in some ways, embodying the vision of JARVIS from Iron Man.


Early GUI agents \cite{deng2024mind2web, yang2023set, zheng2024gpt} were often built on structured inputs, such as HTML pages and Accessibility Trees, interacting with interfaces by invoking predefined system functions. However, this approach has significant limitations. On one hand, it relies heavily on the quality and completeness of underlying metadata, making it insufficient for many application scenarios—particularly when handling GUI data across different platforms \cite{lin2024showui, cheng2024seeclick}. On the other hand, since user interactions primarily occur through the visual interface, these methods lack precise visual perception and interaction capabilities for GUI elements \cite{hong2024cogagent, yang2024aria}. Consequently, vision-based multimodal models \cite{SeeClick, uground, lu2024omniparser, wu2024atlas} have emerged as a powerful alternative to traditional GUI automation methods.
    
While the visual and language understanding capabilities of modern LVLMs~\cite{liu2024visual, zhu2023minigpt, ye2024mplug} provide significant advantages for designing GUI agents, these models still face unique challenges in GUI comprehension. Current LVLMs are primarily trained on general datasets for tasks such as image detection, interpretation, and visual question answering (VQA). However, GUI agents require the ability to accurately understand and locate user interface (UI) elements, highlighting a domain gap between data. To bridge this gap, many studies have focused on constructing specialized GUI datasets \cite{yang2024aria, cheng2024seeclick, sun2024genesis} and fine-tuning LVLMs to improve their element localization capabilities. 

Despite some progress, significant challenges remain in building a GUI agent, particularly for computer use scenarios: 
\textbf {(1) Lack of desktop/computer use data.}
Previous work has primarily focused on collecting screenshots from web and mobile applications, leading to a severe shortage of large-scale, open-source desktop GUI data. The only available large-scale desktop GUI dataset today is OS-Atlas \cite{wu2024atlas}. However, OS-Atlas generates a vast amount of synthetic data using a data synthesis platform on physical machines, which may not accurately reflect real-world application scenarios. Moreover, OS-Atlas is heavily skewed toward Windows, with over 95\% of the data originating from it, while significantly underrepresenting other operating systems such as macOS and Linux.
\textbf {(2) Lack of an efficient data pipeline and rich annotations.}
While web and mobile data collection can be automated using automated testing tools, such as Playwright\footnote{https://playwright.dev/} and Puppeteer\footnote{https://pptr.dev/}, no such tool currently exists for desktop data collection. As a result, acquiring desktop GUI data still relies primarily on manual annotation, which demands substantial human resources and time. Additionally, most existing desktop datasets primarily contain text-based UI elements but lack annotations for other crucial components such as icons and widgets, which are essential for desktop interactions. Furthermore, there is a shortage of detailed descriptions of interactive UI elements, such as region captions~\cite{li2024ferret}, which has been proven to help improve the performance of GUI agents.
\textbf {(3) Lack of desktop test benchmarks.} Existing benchmarks contain very limited data in the desktop domain. For example, ScreenSpot~\cite{cheng2024seeclick}, the most widely used benchmark, only includes 
fewer than 200 screenshots and 350 instructions for desktop data.

To address these challenges, in this paper, we propose an automated data pipeline, {\bf AutoCaptioner}, which can generate rich and diverse data automatically with minimal reliance on human efforts. Focusing on desktop and computer-use data, we leverage this pipeline to create a novel large-scale, high-quality desktop dataset, {\bf DeskVision}. We further build {\bf DeskVision-Eval}, the largest desktop based test benchmark. Finally, by integrating the collected multimodal datasets, we train a powerful GUI visual understanding model, {\bf GUIExplorer}, which achieves state-of-the-art (SOTA) performance. Specifically, our contributions are:

1. \textbf{AutoCaptioner}: We introduce an innovative automated GUI data pipeline that leverages the strengths of UI detection models and large multimodal models to generate large-scale, diverse, real-world scenario data enriched with detailed and informative region captions, while significantly reducing human effort;


    2. \textbf{DeskVision}: Using AutoCaptioner, we introduce DeskVision, the first large-scale desktop dataset focused on real daily user scenarios. DeskVision consists of {\bf 54,855} images with {\bf 303,622} annotations, balanced across different operating systems (including Windows, macOS, and Linux), and features a well-proportioned mix of text-based and icon-based UI elements, enriched with detailed region captions that describe the text, type, and attributes of each element. To address the lack of sufficient benchmarks for desktop use cases, we also present {\bf DeskVision-Eval}, a carefully curated 5,000 sample subset of DeskVision. DeskVision-Eval is more than {\bf 25} times larger than the most widely used desktop benchmark~\cite{SeeClick} today. This benchmark is designed to capture the diversity of real-world use cases, spanning various OS platforms, UI element types, and interactions. We believe DeskVision and DeskVision-Eval will bridge the data gap in GUI research and significantly benefit the development of future GUI agents. Both datasets will be open-sourced;   

3. \textbf{GUIExplorer}: Using DeskVision, along with existing multimodal datasets, we develop the GUI model, GUIExplorer. Without complex architectural designs, GUIExplorer achieves state-of-the-art (SOTA) performance across multiple benchmarks, demonstrating the effectiveness and generalizability of the DeskVision dataset. Furthermore, ablation studies confirm that DeskVision significantly enhances the performance of LVLMs on GUI-related tasks (22.1\% and 26.1\% for Qwen2-VL and LLaVA-One-Vision respectively), highlighting its value in advancing GUI understanding.


\section{Related Work}
\textbf{GUI Agents.} 
GUI agents have been developed to simulate human-like operations. Early GUI agents \cite{deng2024mind2web, yang2023set, zheng2024gpt} were often built on structured inputs, such as HTML pages and Accessibility Trees, interacting with interfaces by invoking predefined system functions. Some research \cite{kim2024language, zheng2023synapse} focused on navigation tasks using LLMs. With the rise of Large Vision Language Models (LVLMs), GUI agents have started to be developed 
using image inputs. 
Auto-UI combined the BLIP2 \cite{li2023blip} visual encoder with FLAN-Alpaca \cite{ghosal2023flacuna} to create a multimodal solution. CoCoAgent \cite{ma-etal-2024-coco} further improved GUI perception by incorporating detailed element layouts and decomposing action predictions. 
The AppAgent series \cite{li2024appagent} used GPT-4 for Android app exploration, while CogAgent improved navigation on mobile and desktop platforms by integrating higher resolution inputs. SeeClick \cite{cheng2024seeclick} focused on GUI grounding, enabling the agent to locate screen elements. MobileVLM \cite{wu2024mobilevlm} specialized in Chinese app navigation, and Ferret-UI \cite{you2025ferret} introduced ``arbitrary resolution" magnification for better element granularity, extending cross-platform support in Ferret-UI 2 \cite{li2024ferret}. 
Most recent research spans mobile \cite{rawles2024androidworld, nong2024mobileflow}, web \cite{koh2024visualwebarena, zhou2023webarena, lai2024autowebglm}, and desktop \cite{xie2024osworld} platforms, highlighting the growing need for adaptable agents to operate seamlessly across diverse environments.

\textbf{Data for Building GUI Agents.} High-quality GUI data is fundamental for developing high performing GUI agents \cite{wu2024copilot}. The field was pioneered by Rico \cite{deka2017rico}, which introduced sequential GUI data for mobile applications, and MiniWob \cite{shi2017world}, which provided foundational keyboard and mouse operation data for web-based tasks. Subsequently, platform-specific datasets have emerged across mobile \cite{rawles2024androidworld, chai2024amex, lu2024gui}, web \cite{liu2018reinforcement, cheng2024seeclick, chen2024guicourse}, and desktop environments \cite{wu2024atlas}. Recent works have emphasized trajectory datasets that integrate GUI information, instructions, and action sequences \cite{li2024effects, zheng2024agentstudio}. However, three primary challenges persist in GUI data collection. First, existing datasets often lack essential interaction elements. Second, manual curation remains costly and inefficient. Finally, most datasets are restricted to specific platforms, such as web only. 

\section{AutoCaptioner: An Automated Data Pipeline}

\begin{figure*}[h!]
    \centering
    \includegraphics[width=1.0\linewidth]{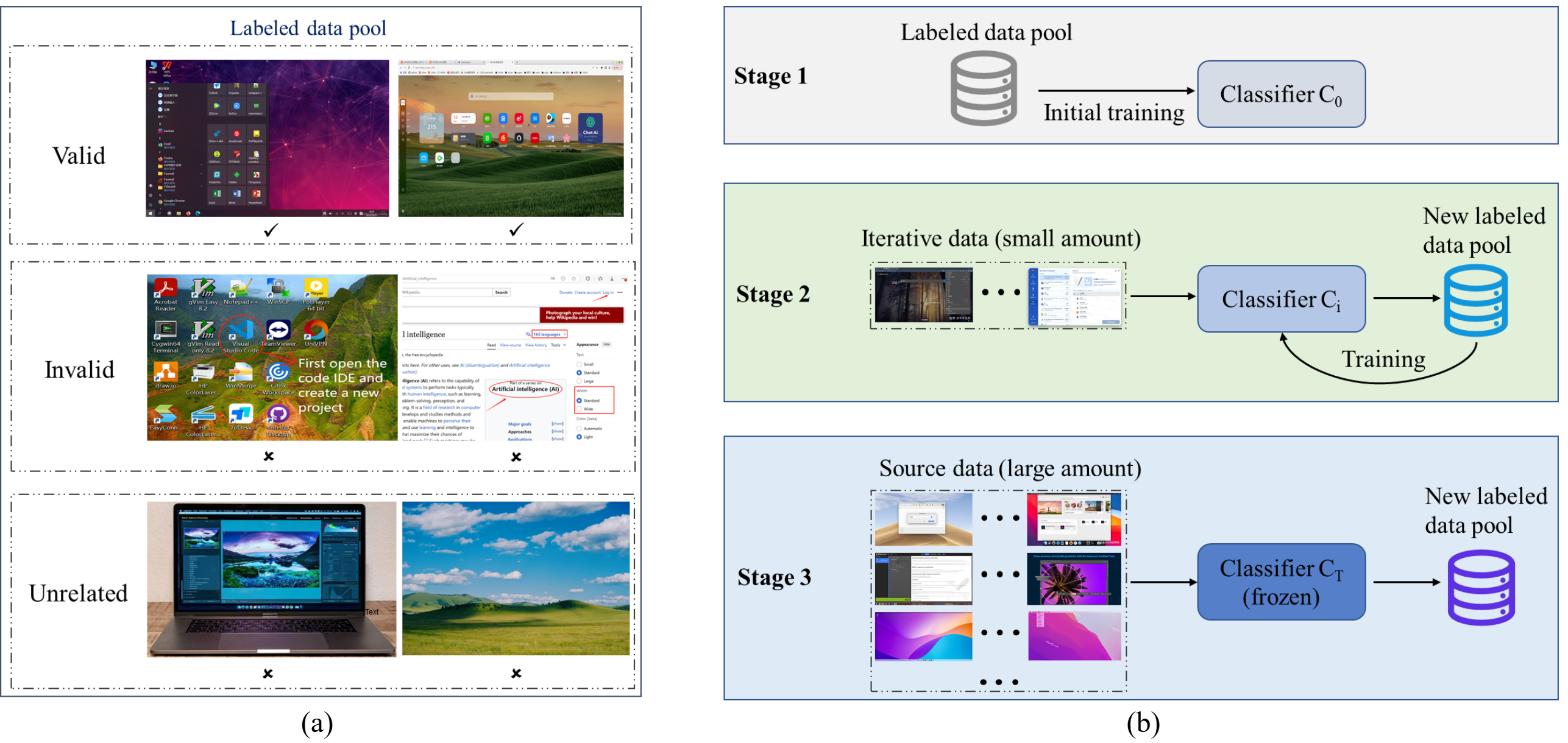}
    \caption{Data Sourcing. (a) Examples of three data types. (b) Our data sourcing pipeline, which consists of three stages. Stage 1 uses limited labeled data to train the initial version of the classifier; Stage 2 employs an iterative training method, sequentially inputting certain numbers of images (5k in our setting) to update the data pool and using the updated pool to iteratively train the classifier. Stage 3 uses the final frozen classifier to clean large amounts of source data.}
    \label{fig:data_sourcing}
\vspace{-0.2in}
\end{figure*}

In this section, we introduce {\bf AutoCaptioner}, an automated data pipeline designed to generate data automatically. AutoCaptioner consists of two key stages: {\bf data sourcing} and {\bf data annotation}. While AutoCaptioner can be applied to generate various types of data, this paper specifically focuses on generating desktop data. The following sections provide a detailed overview of each stage.

\begin{table}[h!]
    \centering
    \tabcolsep=0.06cm
    \begin{tabular}{lccc}
    \hline
       \textbf{Dataset} & \textbf{Desktop} & \textbf{Open Source} & \textbf{Input Text} \\
    \hline
       Ferret-ui & {\textcolor{red}{\ding{53}}} & {\textcolor{red}{\ding{53}}} & Human Ann \\
       CogAgent & {\textcolor{red}{\ding{53}}} & {\textcolor{red}{\ding{53}}} & HTML Text  \\
       GUICourse & {\textcolor{red}{\ding{53}}} & {\textcolor{green}{\ding{51}}} & HTML Text \\
       SeeClick & {\textcolor{red}{\ding{53}}} & {\textcolor{green}{\ding{51}}} & HTML Text \\
       OS-Atlas & {\textcolor{green}{\ding{51}}} & {\textcolor{green}{\ding{51}}} & Element Type \& Name\\
    \hline
       \textbf{DeskVision} & {\textcolor{green}{\ding{51}}} & {\textcolor{green}{\ding{51}}} &  \textbf{Diversified UI Caption} \\
    \end{tabular}
    \caption{DeskVision compared to existing datasets.}
    \label{tab:collections}
    \vspace{-0.2in}
\end{table}

\subsection{Data Sourcing}


As shown in \cref{tab:collections}, while we have a substantial amount of mobile and web data \cite{wu2024atlas} \cite{SeeClick}, we currently lack sufficient data on the desktop side. 
The only accessible desktop dataset at scale is from OS-Atlas~\cite{wu2024atlas}, which uses physical machines and the A11y tree to simulate human interaction for desktop screen data generation. However, the A11y tree focuses solely on executing actions rather than understanding their meaning. Consequently, the generated desktop data may include uncommon functions or software interfaces that do not accurately reflect real-world use cases. Additionally, the OS-Atlas dataset is heavily skewed toward Windows, with over 95\% of the data originating from it, while significantly lacking representation for other operating systems such as macOS and Linux.

\begin{figure*}[h!]
    \centering
    \includegraphics[width=0.95\linewidth]{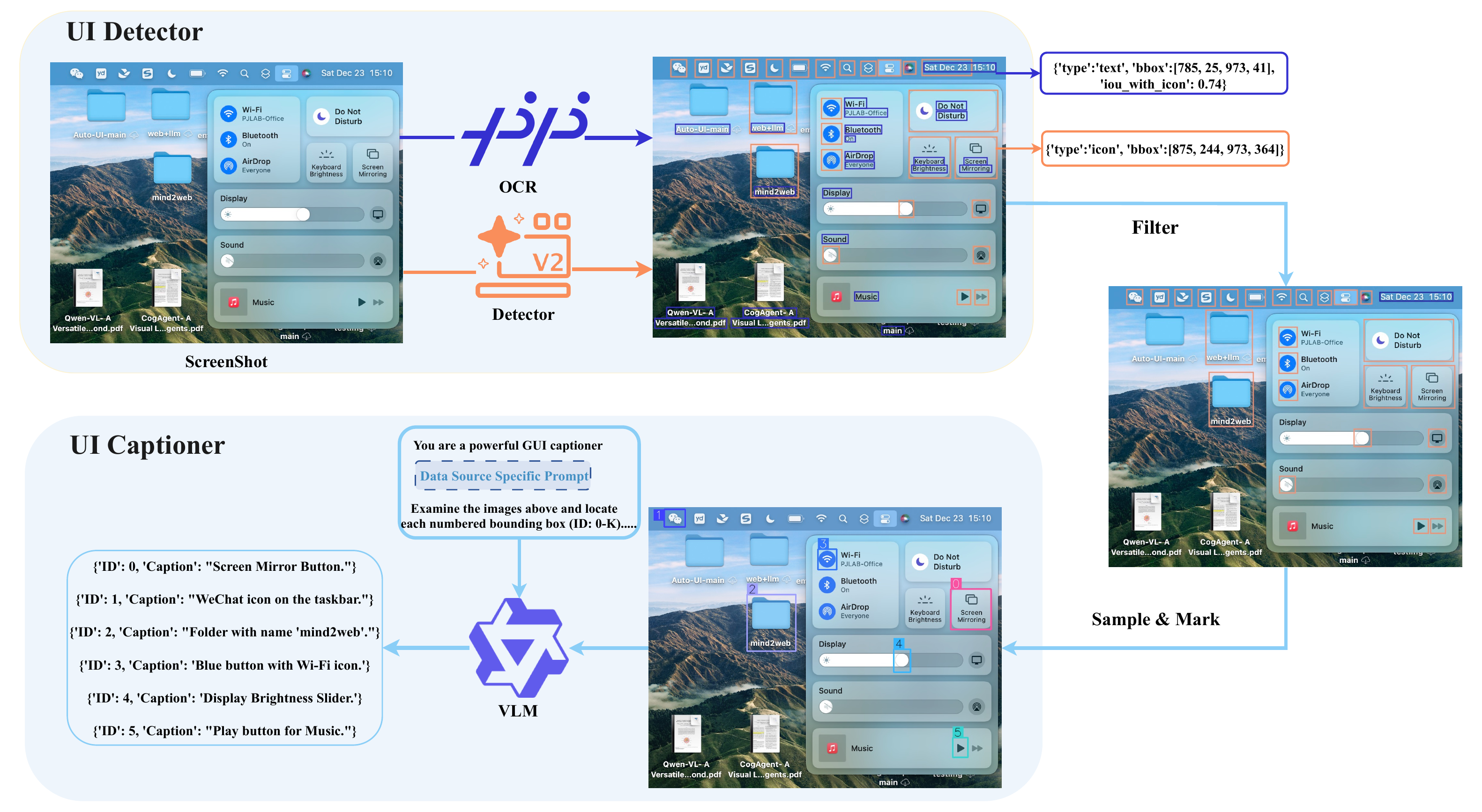}
    \caption{Data annotation pipeline. The screenshot is first sent to UI Detector to detect the interactive UI elements. Detected UI elements are then filtered, sampled and marked. After that, UI elements are sent to UI Captioner to generate final region captions.}
    \label{fig:MarkCap}
    \vspace{-0.1in}
\end{figure*}

To obtain more diverse desktop data that also better reflects daily usage, we begin by sourcing a large volume of desktop screenshot data from the internet. Specifically, we manually compiled a list of 150 common apps and collected screenshots of these apps across three different operating systems: Windows, MacOS, and Linux. 
However, the internet data is often noisy, containing numerous images that fail to meet the required standards. There are two main types of the unqualified data: (1) Unrelated data – Images unrelated to desktop screens, such as landscapes, profile pictures, or product photos; (2) Invalid data – Images relevant to desktop screens but containing additional elements such as text, boxes, arrows, or other annotations, which could interfere with future labeling.

To clean up the data, we build a data filtering pipeline with a classifier in three stages, as shown in \cref{fig:data_sourcing}. In Stage 1, we manually source 5,000 samples for each category (i.e., unrelated, invalid, and valid data), and train the initial classifier. In Stage 2, we use the trained classifier to categorize new data into these three classes. Only data with a confidence score above 0.8 is retained, while the rest is discarded. The retained data is then verified by human labelers and added to the data pool to further refine the classifier. We iterate on this process multiple rounds, improving accuracy and monitoring classification performance. Once the model achieves an accuracy above 95\%, we freeze it. In Stage 3, the final model is used to automatically classify the remaining images, keeping only those with a confidence score of at least 0.9. Ultimately, 
the valid screen data samples will be sent for data annotation. Since invalid data often appear in small interactive areas, we adopt YOLOv5~\cite{yolov5} as our backbone for the classifier, leveraging its detection capabilities to improve the accuracy of data filtering.

\subsection{Data Annotation}

\begin{figure*}[h!]
    \centering
    \includegraphics[width=0.95\linewidth]{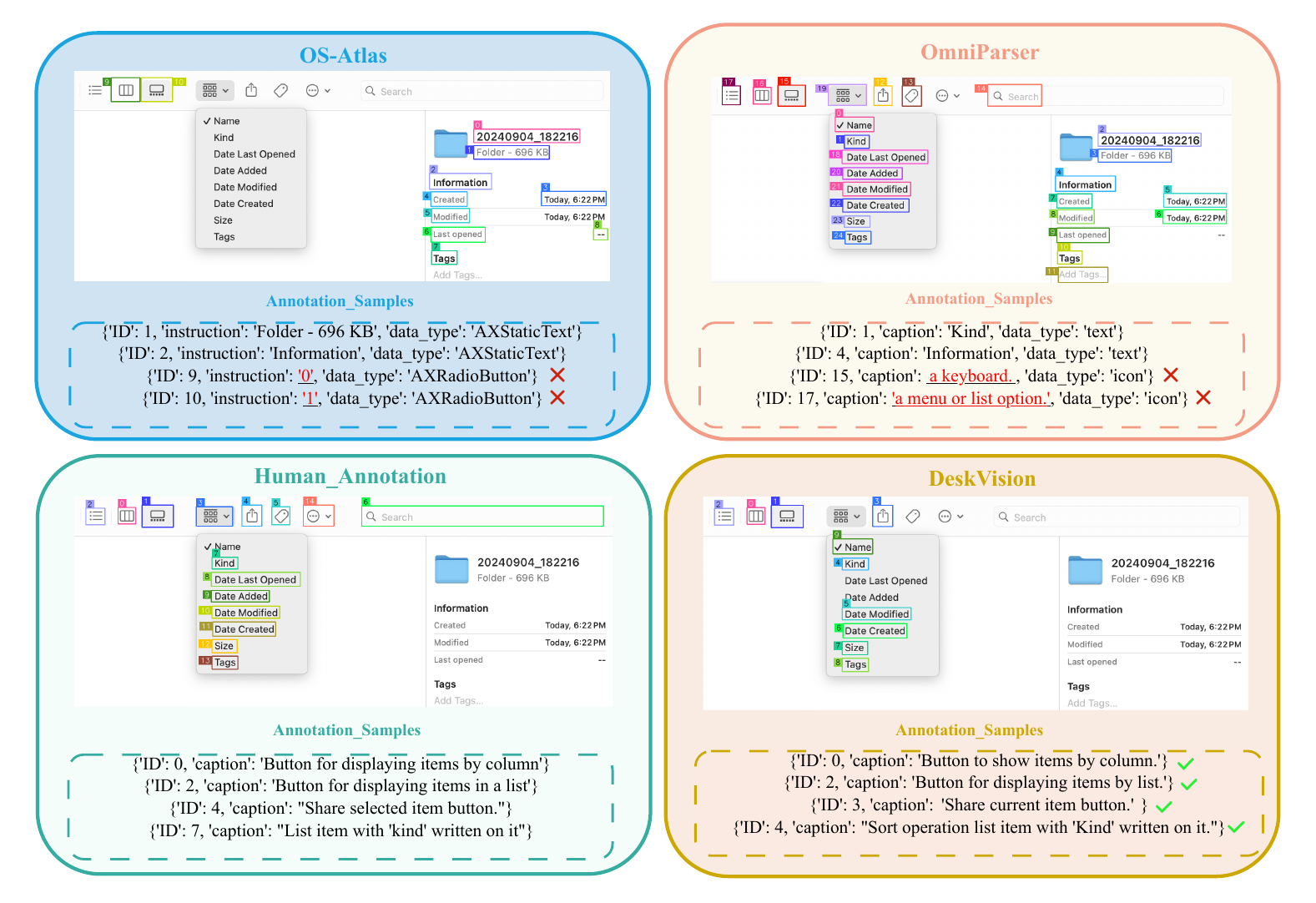}
    \caption{Examples of data annotations on a single screenshot from different methods/datasets. Human\_Annotation's result is manually labeled. Errors in the captions are marked in red, while detailed and accurate captions are marked in green. 
    }
    \label{fig:Qualitative_Analysis}
    \vspace{-0.2in}
\end{figure*}

For GUI-related tasks, given a screenshot, we need to annotate the interactive UI elements within it. The most common interactive elements are interactive text and icons/widgets. For each element, we annotate its bounding box along with a corresponding label. Recent work~\cite{li2024ferret,lin2024showui} demonstrates that incorporating {\bf region captions}, the detailed descriptions of interactive UI elements that cover their visual features, textual content, and functional information, can help GUI agent models more accurately recognize, locate, and understand the role of these elements. However, manually annotating region captions of all elements is labor-intensive and costly. As a result, existing desktop data annotations~\cite{wu2024atlas} primarily focus on visible text on the screen, lacking richer descriptions for icons and widgets, as shown in \cref{fig:Qualitative_Analysis}. On the other hand, today's UI detection models~\cite{lu2024omniparser} exhibit strong UI element localization capabilities but often fail to generate high-quality descriptions for UI elements. Conversely, large visual language models (LVLMs), leveraging their advanced visual understanding, can produce detailed and accurate UI descriptions. However, even SOTA LVLMs (such as GPT-4o and Qwen2-VL-72B) struggle with accurately localizing UI elements. To address this limitation, we propose an automated pipeline 
 that consists of two stages with two key modules: the {\bf UI Detector} and the {\bf UI Captioner}. The UI Detector is responsible for localizing GUI-related elements, while the UI Captioner generates rich descriptions for the detected UI elements, as illustrated in \cref{fig:MarkCap}.  
It is worth noting that while this paper primarily focuses on desktop data, our pipeline is not limited to this domain and can be widely applied to other types of data, such as data from mobile and web interfaces. Below, we describe each module in detail.

\subsubsection{UI Detector}
Given a desktop screenshot as input, the UI Detector identifies the bounding box coordinates of interactive UI elements. We annotate only interactive elements, as they are crucial for the GUI agent. These elements fall into two categories: texts and icons/widgets. To ensure high detection quality, our UI Detector leverages the combined outputs of two models: OmniParser~\cite{lu2024omniparser} and PaddleOCR 
\footnote{https://github.com/PaddlePaddle/PaddleOCR}. OmniParser excels at detecting icons/widgets, while PaddleOCR is highly accurate in text detection. Specifically: (1) If detected text is fully contained within a detected icon, we retain the icon’s bounding box as the output, as there is a high chance that it's a UI element with both texts and icons; (2) If detected text significantly overlaps with an icon (e.g., IoU $\ge$ 0.7), but is not fully contained, most likely it's a text UI element, we retain the text's bounding box due to PaddleOCR's higher text detection accuracy; (3) If detected text has an IoU $\le$ 0.7 with all icons, we discard it, as it is likely non-interactive. We set a high criteria to ensure top-quality annotation. Additionally, we filter out text bounding boxes exceeding a certain width threshold, as long texts are unlikely to be interactive.

\subsubsection{UI Captioner}
To prevent overlapping and overcrowded annotations while increasing data diversity, we sample only a subset of UI elements from each screenshot. Specifically, we first randomly select one UI element as the starting point and compute its Euclidean distance to all other elements. Then, from the five UI elements farthest from the starting point, one is randomly chosen. This process is repeated for 5–8 cycles to ensure that the sampled elements are well distributed across the screen. After sampling, the selected UI elements are sequentially marked with IDs starting from 1, 2, 3, and so on, directly on the original screenshot. We have found that assigning IDs in this manner and passing them along with bounding boxes into the LVLMs significantly improves captioning quality, especially for small objects such as UI elements. An example of the annotated screenshot is shown in \cref{fig:MarkCap}.
 
 Once the interactive UI elements are detected, filtered, sampled and marked, we enrich them with detailed captions. To achieve this, we leverage the strong captioning capabilities of LVLMs. Specifically, we curated a UI dataset by selecting 10,000 web samples from wave-ui\footnote{https://huggingface.co/datasets/agentsea/wave-ui} and manually annotating 500 desktop samples. Despite domain gaps, our experiments show that incorporating web data enhances LVLMs' performance in GUI captioning. We use this dataset to fine-tune Qwen2.5-VL as our final UI Captioner. 
Compared to other datasets, our UI Captioner generates richer descriptions for each labeled UI element in the image, including not only the element's category (e.g., button, icon) but also the text inside it. For instance, as shown in \cref{fig:MarkCap}, the caption for the Wi-Fi icon is `Blue button with Wi-Fi icon'. These detailed descriptions help the model better understand the visual features and functionality of the UI elements. 


\begin{figure*}[h!]
    \centering
    \begin{subfigure}[b]{0.24\textwidth}
        \centering
        \includegraphics[width=\textwidth]{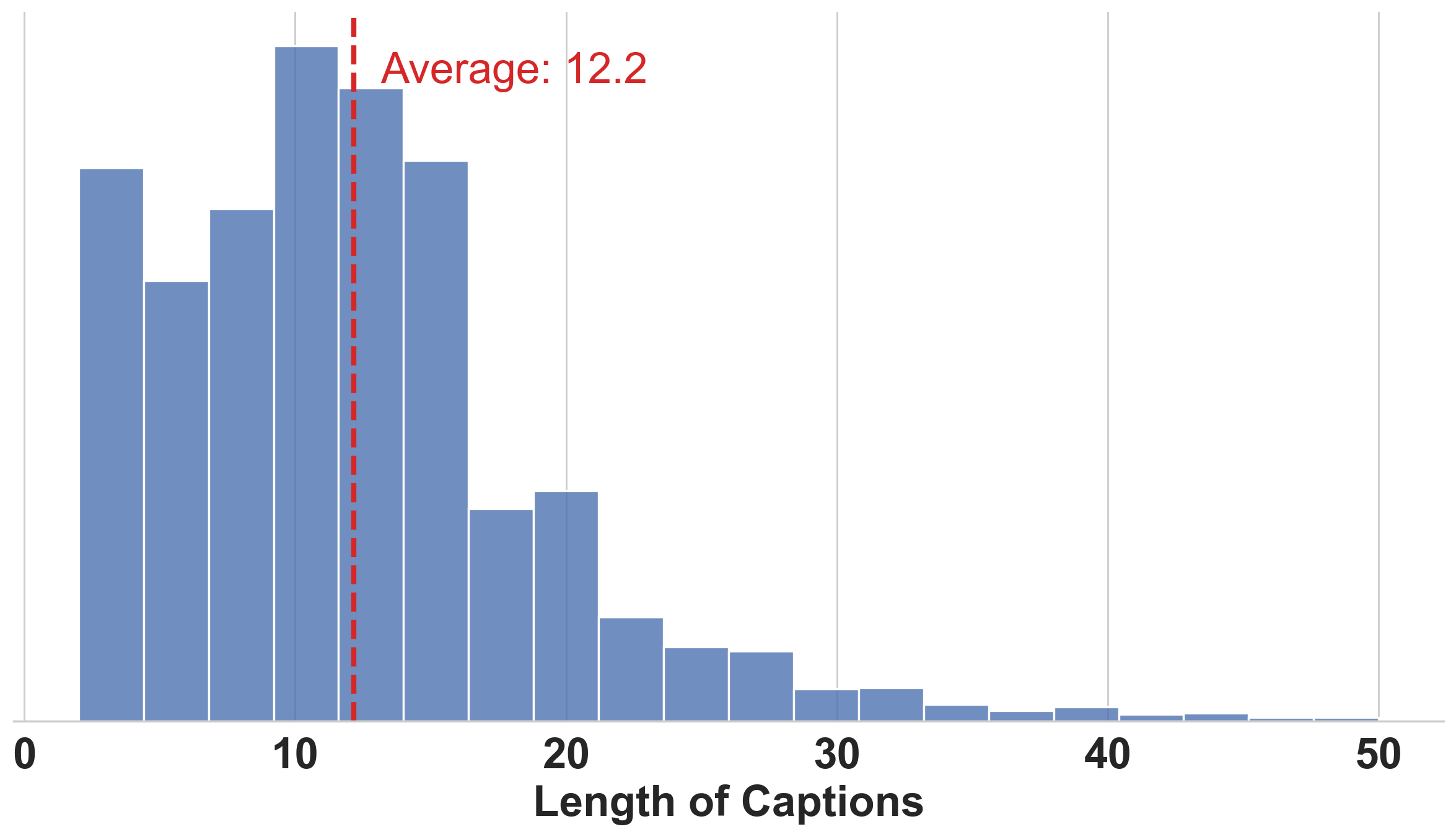} 
        \caption{}
        \label{fig:subfigure_a}
    \end{subfigure}
    \begin{subfigure}[b]{0.24\textwidth}
        \centering
        \includegraphics[width=\textwidth]{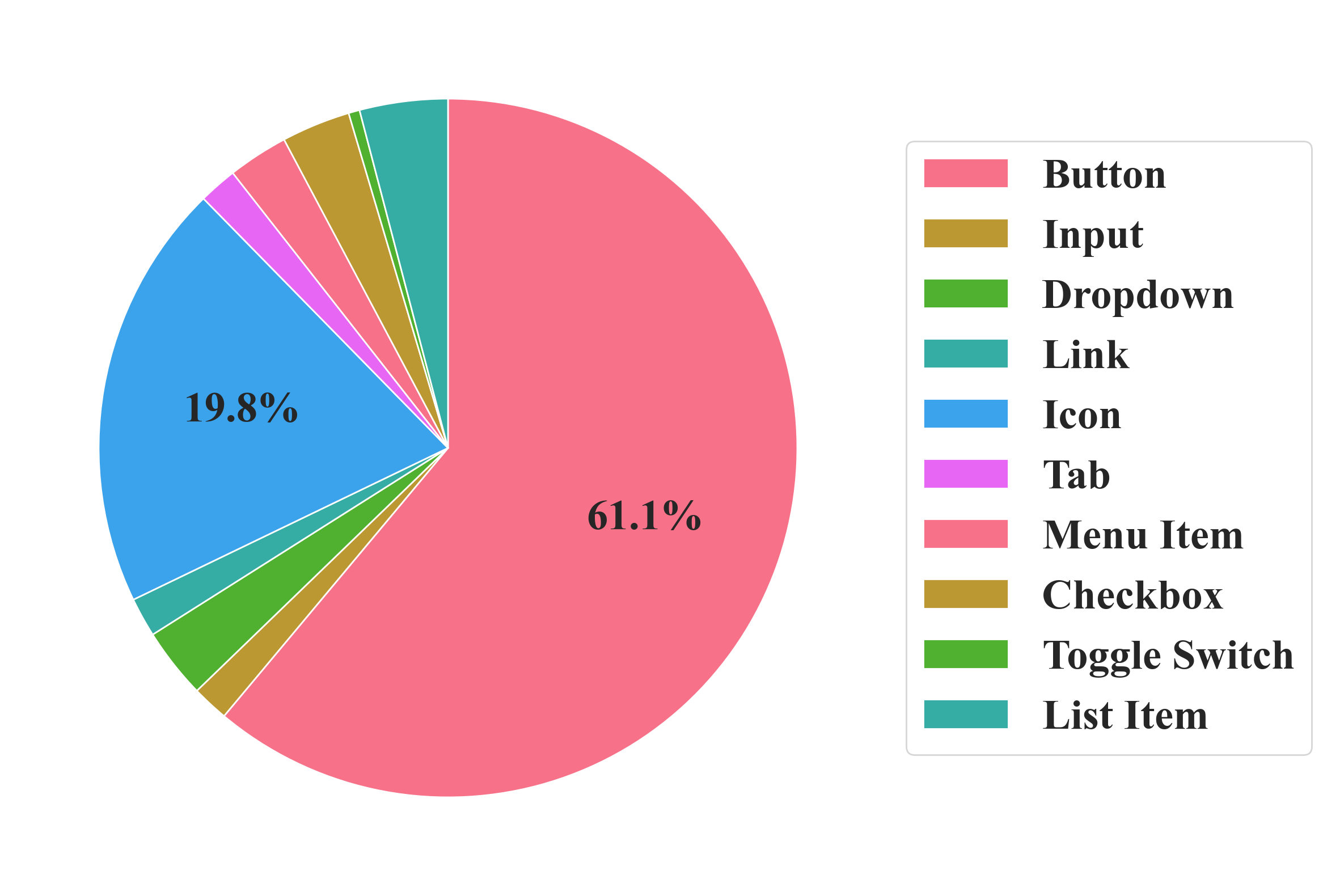} 
        \caption{}
        \label{fig:subfigure_b}
    \end{subfigure}
    \begin{subfigure}[b]{0.24\textwidth}
        \centering
        \includegraphics[width=\textwidth]{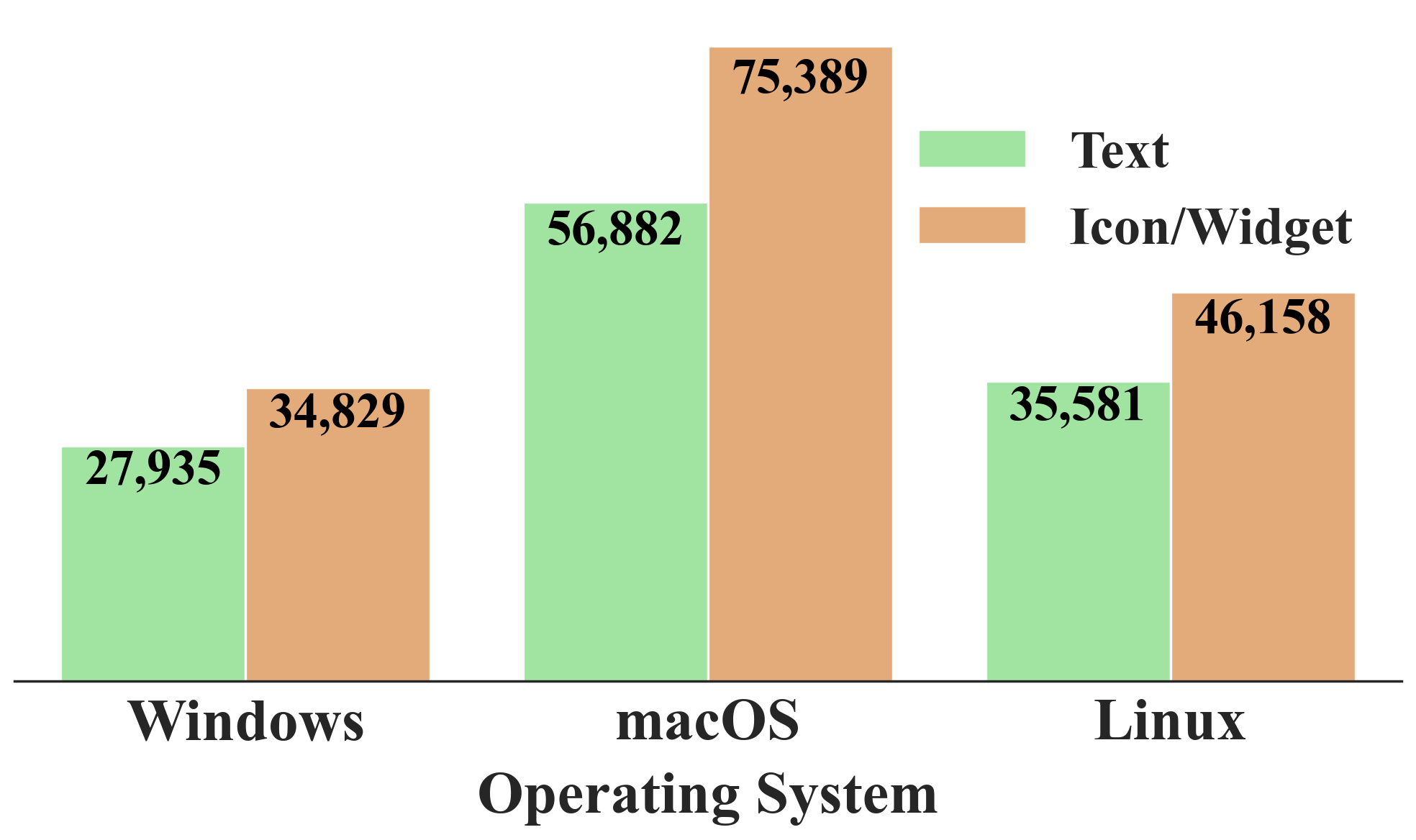} 
        \caption{}
        \label{fig:subfigure_c}
    \end{subfigure}
    \begin{subfigure}[b]{0.24\textwidth}
        \centering
        \includegraphics[width=\textwidth]{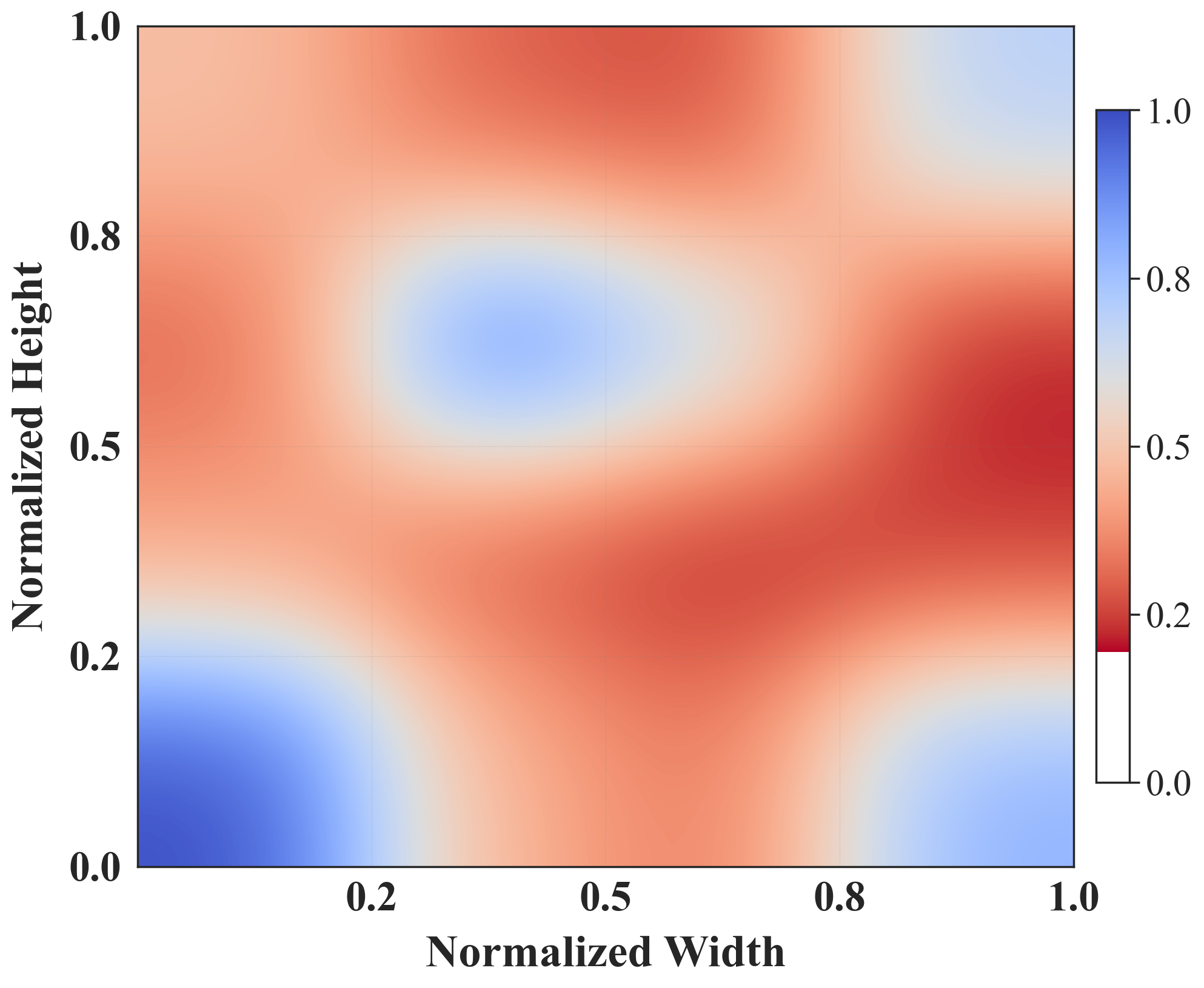} 
        \caption{}
        \label{fig:subfigure_d}
    \end{subfigure}
    
    \caption{Statistics of DeskVision. (a) The distribution of caption lengths. (b) The UI categories of the labelled elements. (c) The types of elements within each OS. (d) The heatmap of the spatial distribution of annotation elements in the normalized image.}
    \label{fig:Quantitative_Analysis}
    \vspace{-0.2in}
\end{figure*}

    
    
    

\section{DeskVision Dataset}
\label{sec:dv}

Using the AutoCaptioner pipeline, we create DeskVision, a new large-scale desktop GUI dataset. DeskVision comprises 54,855 desktop screens with 303,622 annotations, providing a more accurate reflection of daily usage and a well-balanced representation of different operating systems, including Windows, MacOS, and Linux. DeskVision features a well-proportioned mix of text-based and icon/widget-based UI elements, enriched with detailed region captions that describe the text, type, and attribute of each element. Below, we present qualitative and quantitative analysis of it.


\textbf{Qualitative Analysis}. We compare annotations from OS-Atlas \cite{wu2024atlas}, Omniparser \cite{lu2024omniparser}, human labeling, and DeskVision in ~\cref{fig:Qualitative_Analysis}. DeskVision annotations focus more on interactive areas rather than non-interactive static text compared to both OS-Atlas and Omniparser, thanks to our filtering strategy. Additionally, the captions provide richer descriptions of the elements. Notably, the annotations automatically generated by our AutoCaptioner pipeline closely resemble those produced through manual annotation, demonstrating the high accuracy of AutoCaptioner.

\textbf{Quantitative Analysis}. We conduct a statistical analysis of DeskVision, as shown in \cref{fig:Quantitative_Analysis}. First, we examine the distribution of caption lengths in \cref{fig:Quantitative_Analysis}(a). The caption lengths range from 5 to 20 characters, with an average length of 12.2 characters. Next, \cref{fig:Quantitative_Analysis}(b) illustrates the distribution of different UI types in the captions. The Button type is the most prevalent, accounting for 61.1\%, followed by Icon at 19.8\%. Other types, such as Input and Dropdown, appear less frequently but still constitute a notable portion.
\cref{fig:Quantitative_Analysis}(c) presents the distribution of caption information categories across different operating systems, with texts and icons/widgets analyzed separately. The figure shows that their quantities are similar, with icons/widgets being slightly more prevalent. Unlike OS-Atlas, which is highly imbalanced, our dataset is well-balanced across different operating systems, making it a valuable complementary resource.
Finally, we explore the spatial distribution of all annotated UI elements using a normalized heatmap. As shown in \cref{fig:Quantitative_Analysis}(d), the annotated elements are widely distributed, with a slightly higher concentration in the center and at the four corners.

\textbf{DeskVision-Eval}. Existing GUI grounding benchmarks~\cite{SeeClick, chen2024guicourse} contain only a limited amount of desktop data. For example, ScreenSpot~\cite{SeeClick} includes approximately 600 screenshots and 1,200 instructions, with fewer than 200 screenshots and 350 instructions specifically for desktop data. Similarly, GUI-Env~\cite{chen2024guicourse} comprises 1,877 instructions but primarily evaluates grounding and OCR capabilities in the web domain, with no desktop data. To better evaluate the capability of GUI models in understanding desktop scenarios, we introduce DeskVision-Eval, a carefully curated 5,000 samples from the DeskVision dataset to serve as a test benchmark. These samples capture the diversity of real-world use cases, spanning various operating system platforms, UI element types, and interactions. To the best of our knowledge, DeskVision-Eval is the largest open sourced desktop test benchmark currently.


\section{Experiments}
In this section, we first introduce our GUI understanding model, GUIExplorer, followed by its training details. Next, we present the benchmarks and evaluation metrics used to assess GUIExplorer. We then compare the performance of GUIExplorer to the SOTA models. Finally, to analyze the effectiveness of the DeskVision dataset, we compare multiple LVLMs trained with DeskVision.

\subsection{Model}
Our model uses a similar architecture as LLaVA-OneVision \cite{llava-ov}, where a visual encoder is built to process the screenshot data, the LLM handles the text data (including both input and output text), and the projector aligns the visual and textual embeddings. We select SigLip \cite{siglip} as our visual encoder. Compared to CLIP \cite{clip}, SigLip performs better with smaller batch sizes, making it more suitable for GUI tasks, which often involve larger images and inherently small batch sizes during training. We use Qwen-2 \cite{qwen2} as our base LLM. A two-layer MLP is chosen as our projector to align the visual features to the text embedding space.

\begin{table*}[]
\centering
\resizebox{1.0\linewidth}{!}{
\fontsize{8}{10}\selectfont
\begin{tabular}{cccccccc}
\hline
\multirow{2}{*}{Models} & \multicolumn{2}{l}{Mobile}     & \multicolumn{2}{l}{Desktop}     & \multicolumn{2}{l}{Web}         & \multirow{2}{*}{Avg} \\ \cline{2-7}
                                  & Text          & Icon/Widget    & Text           & Icon/Widget    & Text           & Icon/Widget    &                      \\ \hline
Fuyu \cite{fuyu}                           & 41            & 1.3            & 33             & 3.6            & 33.9           & 4.4            & 19.5                 \\ 
CogAgent \cite{hong2024cogagent}                         & 67            & 24             & 74             & 20             & 70.4           & 28.6           & 47.4                 \\ 
SeeClick \cite{cheng2024seeclick}                         & 78            & 52             & 72.2           & 30             & 55.7           & 32.5           & 53.4                 \\ 
InternVL-2-4B \cite{internvl}                    & 9.16          & 4.8            & 4.64           & 4.29           & 0.87           & 0.1            & 4.32                 \\ 
Qwen2-VL-7B \cite{qwen2}                     & 61.34         & 39.29          & 51.01          & 44.98          & 33.04          & 21.84          & 42.89                \\ 
UGround-7B \cite{uground}                       & 82.8          & 60.3           & 82.5           & 63.6           & 80.4           & 70.4           & 73.3                 \\ 
OminiParser(w. LS+ID) \cite{lu2024omniparser}            & \textbf{93.9} & 57             & 91.3           & 63.6           & 81.3           & 51             & 73                   \\ 
OS-Atlas-Base-7B \cite{wu2024atlas}                 & 93.04          & 72.93          & \textbf{91.75} & 62.86          & \textbf{90.87} & 74.27          & 82.47       \\ 
\rowcolor{gray!10}
\textbf{GUIExplorer}  & 89.01 & \textbf{77.29} & 88.14 & \textbf{75.0} & 82.61 & \textbf{81.55} & \textbf{82.86}   \\ \hline
\end{tabular}
}
\caption{Performance of different LVLMs on ScreenSpot Benchmark. Metric is Element Accuracy.}
\label{tab:ScreenSpot}
\end{table*}

\begin{table*}[]
\resizebox{1.0\linewidth}{!}{
\fontsize{8}{10}\selectfont
\centering
\begin{tabular}{ccccccc}
\hline
\multicolumn{1}{c}{\multirow{2}{*}{Models}} & \multicolumn{2}{c}{OCR}                 & \multicolumn{4}{c}{Grounding}                                    \\ \cline{2-7} 
\multicolumn{1}{c}{}                                  & EM Score       & \multicolumn{1}{l}{F1 Score} & IoU@0.2        & IoU@0.5        & IoU@0.7       & Element Accuracy    \\ \hline
SeeClick \cite{cheng2024seeclick}                                              & 5.19 & 8.59	& 53.34	& 24.58	& 5.55	& 56.85          \\
GUICourse \cite{chen2024guicourse} & 44.12 & 64.78	& 68.02	& 47.96	& 23.28	& - \\
UGround-7B \cite{uground}                                            & -              & -                             & -              & -              & -             & 63.76          \\
OS-Atlas-Base-7B \cite{wu2024atlas}                & 42.33 & 60.51 & 76.33 & 59.68 & 41.9	& 75.76       \\ 
\rowcolor{gray!10}
\textbf{GUIExplorer}                       & \textbf{54.60}      & \textbf{78.71}                     & \textbf{88.51}      & \textbf{82.56}      & \textbf{62.17}     & \textbf{87.66}      \\ \hline
\end{tabular}
}
\caption{Performance of different LVLMs on GUI-Env Benchmark. }
\label{tab:GUIEnv}
\vspace{-0.2in}
\end{table*}

\begin{table*}[]
\centering
\fontsize{11}{12}\selectfont
\begin{tabular}{ccccc}
\hline
\multicolumn{1}{c|}{\multirow{2}{*}{Models}}               & \multicolumn{4}{c}{Grounding}                                    \\ \cline{2-5} 
\multicolumn{1}{c|}{}                                   & IoU@0.2        & IoU@0.5        & IoU@0.7       & Element Accuracy    \\ \hline
SeeClick \cite{cheng2024seeclick}                                              & 10.44          & 2.51     & 0.42          & 11.90                  \\
UGround-7B \cite{uground}                                                      & -              & -              & -             & 34.24          \\
OS-Atlas-Base-7B \cite{wu2024atlas}                                            & 36.53           & 21.09          & 9.60         & 35.28           \\ 
\rowcolor{gray!10}
\textbf{GUIExplorer}                       & \textbf{73.27}      & \textbf{67.64}      & \textbf{58.45}     & \textbf{72.02}      \\ \hline
\end{tabular}
\caption{Performance of different LVLMs on DeskVision-Eval Benchmark.}
\label{tab:DeskVision-Benchmark}
\end{table*}

\begin{table*}[h!]
\centering
\resizebox{1.0\linewidth}{!}{
\fontsize{8}{10}\selectfont
\begin{tabular}{cc|cccc}
\hline
\textbf{Models} & \textbf{Strategies} & \textbf{Mobile} & \textbf{Desktop} & \textbf{Web} & \textbf{Avg} \\ \hline
\multirow{5}{*}{Qwen2-VL-7B\cite{qwen2}} & baseline & 39.29 & 44.98	& 21.84	& 35.37 \\
& OS-Atlas-desktop & 61.57 & 47.14	& 64.08	& 57.6 \\ 
& OS-Atlas-desktop + DeskVision & \textbf{71.25} & \textbf{68.22} & \textbf{71.55} & \textbf{70.34}    \\   \hline
\multirow{5}{*}{LLaVA-OneVision-7B\cite{llava-ov}} & baseline & 0 & 0.7 & 0.5 & 0.4 \\
& OS-Atlas-desktop & 8.3 & 15.71 & 51.94 & 25.32 \\ 
& OS-Atlas-desktop + DeskVision & \textbf{15.82} & \textbf{25.86} & \textbf{54.1} & \textbf{31.92} \\
\hline
\end{tabular}
}
\caption{
Performance of DeskVision data across various LVLMs on ScreenSpot (Icon/Widget) Benchmark. OS-Atlas-desktop represents the desktop data from OS-Atlas, OS-Atlas-desktop + DeskVision represents combining two datasets together.
}
\label{tab:DeskVision}
\vspace{-0.1in}
\end{table*}

\subsection{Data Construction}
We integrate multiple public datasets with our DeskVision dataset (excluding DeskVision-Eval) for training, encompassing mobile, web, and desktop UI scenarios. 

\textbf{Mobile UI Data.}
We follow the OS-Atlas to use the mobile data from four main sources: 
AMEX, UIBert, Widget Captioning, and RICOSCA. The Android Multi-annotation EXpo (AMEX) is a comprehensive, large-scale dataset designed for generalist mobile GUI-control agents \cite{chai2024amex}. UIBert \cite{uibert} is an extension of the Rico dataset \cite{deka2017rico}, created for two tasks: similar UI component retrieval and referring expression component retrieval. Widget Captioning data were collected by \cite{widget}. RICOSCA is a dataset automatically labeled using Android VH in \cite{mapping}. These datasets primarily provide the following annotation information: widget captioning, mobile UI grounding, and mobile UI summarization.

\textbf{Web UI Data.}
Existing work has made available a considerable amount of web data, such as OS-Atlas \cite{wu2024atlas}, which provides 1.6 million webpage screenshots and 7.7 million elements. Due to the resource limit, We use only a randomly sampled 40\% subset of the data used by OS-Atlas to train our model. 

\textbf{Desktop UI Data.} For public datasets, OS-Atlas \cite{wu2024atlas} generates desktop data by using the data synthesis platform and simulation environment. It contains about 12,000 screenshots, including 267 screenshots in Linux, 348 screenshots in MacOS, and 11,768 screenshots in Windows. In addition to the public datasets, we also train our model using DeskVision (training split only).

\textbf{General Data.}
To equip our model with essential visual understanding capabilities, we used the VQA and Visual Reasoning SFT data collected by LLaVA-OneVision, totaling 1.36 million instruction-labeled single-image (800K) and multi-image (560K) data.

\subsection{Training Details}
During training, we activate all model modules, including the visual encoder, LLM, and MLP projector. The training data includes a mix of sampled general data, fully collected mobile, desktop, and web data, as well as DeskVision data. We train the model with arbitrary resolution. Initially, we pre-train the model using the general data for the grounding task, providing the model with foundational visual understanding capabilities. Subsequently, we fine-tune the pre-trained model with GUI instruction data and DeskVision data to enhance its ability to understand and execute instructions.

\subsection{Benchmarks and Metrics}
We evaluate our model on two of the most widely used benchmarks for GUI understanding: ScreenSpot \cite{cheng2024seeclick} and GUIEnv \cite{chen2024guicourse}. Additionally, we conduct evaluation on our DeskVision-Eval benchmark.

\textbf{ScreenSpot} \cite{SeeClick} is an evaluation benchmark for GUI grounding, comprising over 600 screenshots and 1,200 instructions from iOS, Android, macOS, Windows and Web environments, along with annotated element types (Text or Icon/Widget). We use the same Element Accuracy metric as \cite{SeeClick} to evaluate the model’s grounding capability. Element Accuracy refers to the accuracy with which the center point of the model's predicted bounding box falls within the ground truth element's bounding box.

\textbf{GUI-Env} ~\cite{chen2024guicourse} is introduced in GUI-Course~\cite{chen2024guicourse} to evaluate the OCR and grounding capabilities of the model on the ``text" elements. Type Exact Match Score (EM Score) and F1 Score~\cite{mrc} are used to evaluate the OCR task, while the accuracy of different IoU values (IoU@*) and Element Accuracy metrics are used to evaluate the grounding task. The EM Score measures the accuracy of predicting action names, without considering the parameters of the actions.  

\textbf{DeskVision-Eval} is described in  Section~\ref{sec:dv}. 
We use Element Accuracy and the accuracy of different IoU values (IoU@*) as evaluation metrics.

\subsection{Results Comparison} 

We compare GUIExplorer to SOTA models on ScreenSpot, GUI-Env, and DeskVision-Eval in \cref{tab:ScreenSpot}, \cref{tab:GUIEnv}, and \cref{tab:DeskVision-Benchmark}, respectively. 
Since there is a significant lack of benchmark results, we run top performing models, e.g., SeekClick~\cite{SeeClick}, UGround-7B~\cite{uground}, and OS-Atlas-Base-7B~\cite{wu2024atlas} using the released inference code to generate results on the GUI-Env and DeskVision-Eval benchmarks. The rest of results are from the original papers. From the results, 
GUIExplorer achieves SOTA performance on both GUI-Env and DeskVision-Eval. It also achieves SOTA performance across all Icon/Widget categories on Screenspot, with an highest average score. 
Since we follow a standard training protocol and dataset, with DeskVision as the only addition, and without using complex architectural designs, the improvement mainly stems from using DeskVision, demonstrating the effectiveness and generalizability of not only GUIExplorer but also the DeskVision dataset. Notably, DeskVision enhances performance not only in desktop scenarios but also on mobile and web platforms, demonstrating its broad applicability and value. However, there is a slight decrease in performance on text-related categories, likely due to our use of a less advanced network architecture compared to others. For instance, OmniParser employs a dedicated OCR module for text, while OS-Atlas uses Qwen2-VL as its backbone, which is more powerful for text detection than Llava-OneVision. 
In GUI tasks, Icon/Widget elements are usually more interactive than Text elements, so improving the grounding capability of these elements will significantly enhance the interactive experience for GUI agents.

\subsection{Ablation Studies on DeskVision}


To further evaluate the effectiveness of the DeskVision data, we test the model fine-tuned with DeskVision against the baseline. We use Qwen2-VL-7B \cite{qwen2} and LLaVA-OneVision-7B \cite{llava-ov} as the baseline LVLM models, and compare the results of the baseline, fine-tuned with OS-Atlas desktop data, and with a combination of OS-Atlas desktop data and our DeskVision data on ScreensSpot in~\cref{tab:DeskVision}. Results show that adding DeskVision data significantly improves the accuracy of both LVLM models (22.1\% and 26.1\% for Qwen2-VL and LLaVA-One-Vision respectively, comparing to using OS-Atlas-desktop only), demonstrating the effectiveness and generalizability of DeskVision.

\section{Conclusion}
In this paper, we have introduced AutoCaptioner, an efficient and automated data pipeline for the creation of GUI data. Using AutoCaptioner, we developed DeskVision, the first large-scale desktop dataset focusing on real-world use case scenarios, along with the largest desktop benchmark DeskVision-Eval. Trained with DeskVision, our GUI model has achieved SOTA performance across multiple benchmarks. We further verified the effectiveness of the DeskVision dataset in understanding visual elements through ablation studies on various LVLMs. We believe DeskVision will help bridge the current data gap and advance the field of GUI agents. Moving forward, we plan to expand DeskVision to include trajectory-based GUI desktop data to facilitate the development of multi-step GUI agents.

    { \small 
    \bibliographystyle{ieeenat_fullname}
    \bibliography{main} }

\end{document}